\documentclass[letterpaper, 10 pt, conference]{ieeeconf}  

\IEEEoverridecommandlockouts                              

\usepackage{cite}
\usepackage{amsmath,amssymb,amsfonts}
\usepackage{algorithmic}
\usepackage{graphicx}
\usepackage{subfigure}
\usepackage{textcomp}
\usepackage{adjustbox}
\usepackage{pifont}
\usepackage{multirow}
\usepackage{fixltx2e}
\usepackage{booktabs}
\usepackage{algorithm}
\usepackage{algorithmic}
\newcommand{\cmark}{\ding{51}}
\newcommand{\xmark}{\ding{55}}
\usepackage{xcolor}
\def\BibTeX{{\rm B\kern-.05em{\sc i\kern-.025em b}\kern-.08em
    T\kern-.1667em\lower.7ex\hbox{E}\kern-.125emX}}

\title{\LARGE \bf
RetinaFaceMask: A Single Stage Face Mask Detector for \\ Assisting Control of the COVID-19 Pandemic
}

\author{Xinqi Fan$^{*}$,
Mingjie Jiang$^{*}$
\thanks{$^{*}$These two authors contributed equally to this paper.}
\thanks{The authors are with the City University of Hong Kong, Hong Kong SAR, China 
        {\tt\small \{xinqi.fan, minjiang5-c\}@my.cityu.edu.hk}. 
        }%
}

\begin{document}

\maketitle
\thispagestyle{empty}
\pagestyle{empty}

\begin{abstract}
    Coronavirus 2019 has made a significant impact on the world. One effective strategy to prevent infection for people is to wear masks in public places. Certain public service providers require clients to use their services only if they properly wear masks. There are, however, only a few research studies on automatic face mask detection. In this paper, we proposed RetinaFaceMask, the first high-performance single stage face mask detector. First, to solve the issue that existing studies did not distinguish between correct and incorrect mask wearing states, we established a new dataset containing these annotations. Second, we proposed a context attention module to focus on learning discriminated features associated with face mask wearing states. Third, we transferred the knowledge from the face detection task, inspired by how humans improve their ability via learning from similar tasks. Ablation studies showed the advantages of the proposed model. Experimental findings on both the public and new datasets demonstrated the state-of-the-art performance of our model. 
\end{abstract}

\section{INTRODUCTION}
According to the World Health Organization (WHO), coronavirus disease 2019 (COVID-19) has infected over 79.2 million individuals and caused over 1.7 million fatalities until the end of 2020~\cite{world2020coronavirus}. Numerous computer-assisted approaches have been developed to aid in the fight against COVID-19, including automatic detection of COVID-19 cases based on X-ray or computed tomography (CT) images~\cite{tabarisaadi2020deep, shamsi2021uncertainty}, COVID-19 trend prediction~\cite{kunjir2020comparative}, and analysis of human reactions to COVID-19~\cite{rafi2020understanding}. It is, however, more critical for individuals to protect themselves from the COVID-19 virus. Fortunately, the study~\cite{cheng2021face} demonstrated that surgical face masks can help limit coronavirus dissemination. At the moment, the WHO recommends that people wear face masks if they have respiratory symptoms or are caring for someone who does~\cite{feng2020rational}. Additionally, several public service providers require users to use services only while wearing masks~\cite{fang2020transmission}. Therefore, automatic face mask detection has emerged as a critical computer vision task for assisting the worldwide community, but research on this is limited.

\begin{figure}[!t]
    \vspace{0.2cm}
    \centering
    \includegraphics[keepaspectratio=true,width=0.25\linewidth, width=20pc]{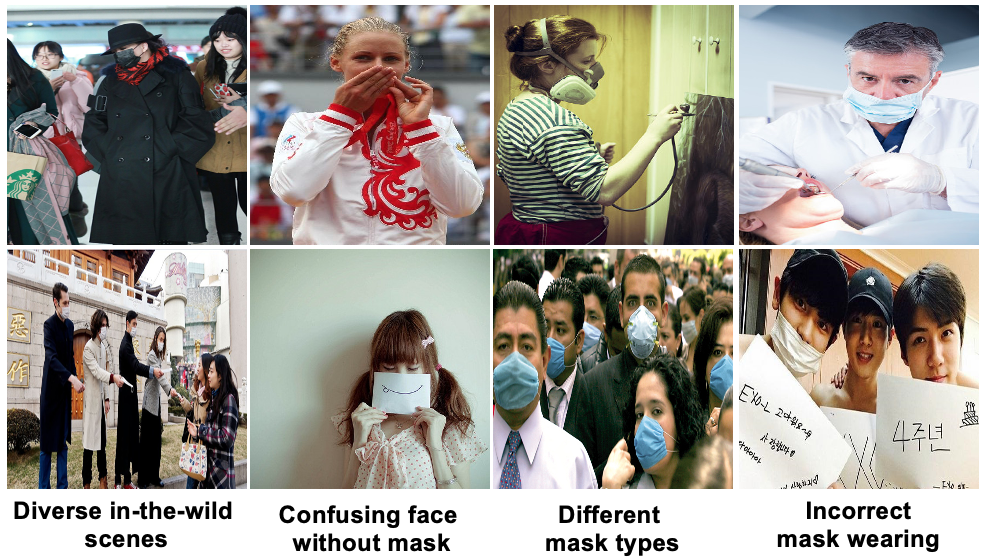}
    \caption{Challenges in face mask detection.}
    \label{fig:dataset_examples}
\end{figure}

Face mask detection entails both the localization of faces and the recognition of mask wearing states, which we define the states as no mask wearing and mask wearing in general. Due to the requirements of healthcare, we further classified the states of mask wearing into correct and incorrect mask wearing states. In one aspect, the face mask detection problem is similar to face detection~\cite{kumar2019face}, as localizing the face is a critical subtask. In another perspective, the problem is closely related to general object detection~\cite{zhao2019object}, where each state can be treated as a distinct class. As shown in Fig.~\ref{fig:dataset_examples}, the challenges of face mask detection include a variety of in-the-wild situations with a complex background, confused faces without masks where faces may be obscured by other objects, a variety of mask types with different shapes and colors, and incorrect mask wearing cases.

Typically, traditional object detectors are built on hand-crafted feature extractors. The Viola Jones detector utilized the Haar feature in conjunction with the integral image approach~\cite{viola2001rapid}, whilst other studies utilized a variety of feature extractors, including the histogram of oriented gradients (HOG), the scale-invariant feature transform (SIFT), and others~\cite{felzenszwalb2008discriminatively}. Recently, object detectors based on deep learning demonstrated superior performance and have dominated the development of new object detectors. Without relying on prior knowledge to construct feature extractors, deep learning can learn the features in an end-to-end manner~\cite{liu2020deep}. There are two types of deep learning based object detectors: one-stage and two-stage detectors. One-stage detectors, such as you only look once (YOLO)~\cite{redmon2016you} and single shot detector (SSD)~\cite{liu2016ssd}, detected objects using a single neural network. The advantage of SSD is that it detects objects using multi-scale feature maps. By contrast, two-stage detectors, such as region-based convolutional neural network (R-CNN)~\cite{girshick2014rich} and faster R-CNN~\cite{ren2015faster}, employed two networks to conduct a coarse-to-fine detection. RetinaFace~\cite{deng2020retinaface}, a dedicated face mask detector, used a multi-scale detection architecture similar to SSD but included a feature pyramid network (FPN) to fuse high and low level semantic information to increase detection performance. Additionally, numerous approaches for studying face mask detection were created. According to the timeline, the initial version of this work, RetinaFaceMask (also known as RetinaMask), can be considered as the first attempt to introduce the face mask detection work. Li \textit{et al.}~\cite{li2020robust} increased the robustness of face mask detection by implementing a mix-up and multi-scale technique based on YOLOv3. To enhance the post-processing of YOLOv3 for face mask detection, a distance intersection over union (IoU) non-maximum suppression (NMS) approach was utilized~\cite{ren2020mask}. However, these algorithms either ignore all possible face mask wearing states that occur in real healthcare applications, or report performance only on limited datasets.

In this paper, we proposed a novel single stage face mask detector, RetinaFaceMask, which is able to detect face masks and contribute to public healthcare. We made the following contributions in this study:
\begin{itemize}
	\item By reannotating the current MAsked FAces (MAFA) dataset used for masked face analysis, we created a new dataset MAsked FAces for Face Mask Detection (MAFA-FMD). The new annotation includes three distinct mask wearing states: no mask wearing, correct mask wearing, and incorrect mask wearing, which is more realistic in terms of contributing to public health. MAFA-FMD contains around 56,000 annotations.
	\item To focus on learning discriminated features associated with face mask wearing states, we proposed a novel context attention module (CAM). The module can extract more useful context features, and concentrate on those that are critical for face mask wearing states. 
	\item Inspired by how humans enhance their skills via the use of knowledge gained from other tasks, we used transfer learning (TL) to transfer the knowledge learned from face detection tasks. Experimentally, we demonstrated that face detection and face mask detection are highly correlated, and the feature learned from the former is useful for the latter task.
\end{itemize}

Ablation studies showed the effectiveness of the CAM and TL, since they can boost the mean average precision (mAP) by a large margin. Experimental results on the public dataset AIZOO demonstrated that RetinaFaceMask achieved the state-of-the-art result, and a $4\%$ increase compared to the baseline method. RetinaFaceMask also had the best performance on the MAFA-FMD dataset, which contains three distinct mask wearing states and is notoriously difficult.

The remainder of this paper is structured as follows. Section II illustrates the established dataset. Section III presents the proposed RetinaFaceMask. Section IV discusses the used datasets, experiment settings, results, and discussion. Finally, Section V concludes the paper and outlines future work.

\begin{figure*}[!h]
    \centering
    \vspace{0.2cm}
    \includegraphics[keepaspectratio=true, width=0.25\linewidth, width=35pc]{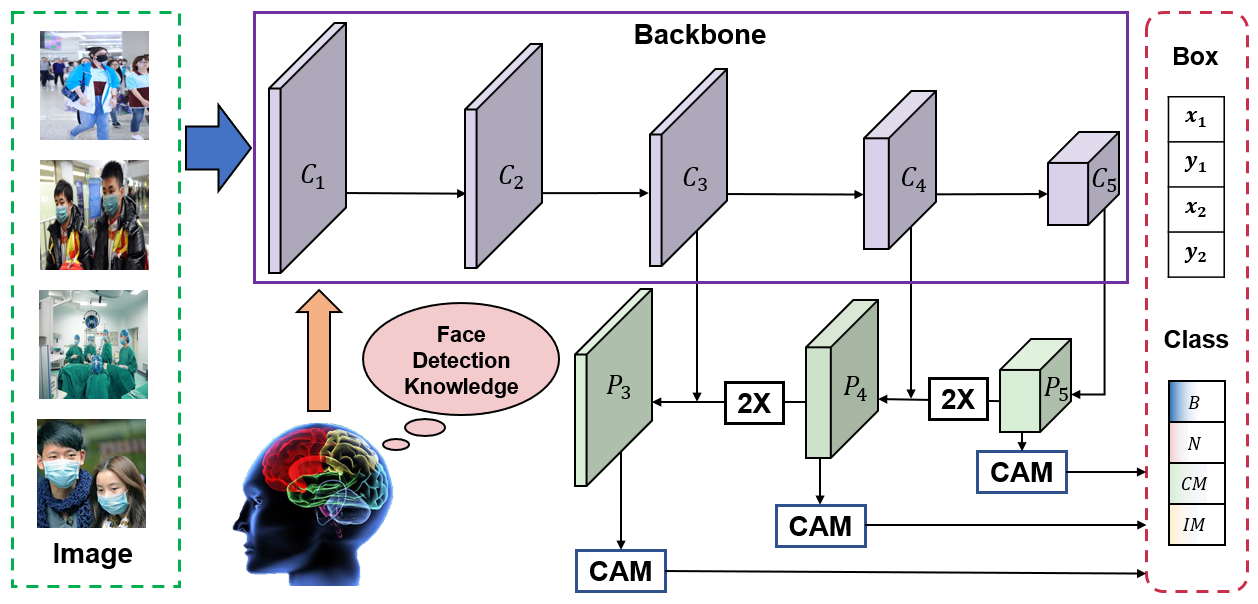}
    \caption{Architecture of RetinaFaceMask. FPN fuses high-level information with low-level information through upsampling and adding; CAMs can focus on learning discriminated face mask wearing states related features; knowledge learned from face detection is transferred into the backbone; $B$, $N$, $CM$ and $IM$ stand for background, no mask wearing, correct mask wearing, and incorrect mask wearing.}
    \label{fig:RetinaFaceMask_arch}
\end{figure*}


\begin{table}[!t]
\vspace{0.5cm}
\label{table:data_diff}
\centering
\caption{Differences between MAFA and MAFA-FMD}
\begin{tabular}{|l|c|c|}
\hline
                                     & \textbf{MAFA}                                                                        & \textbf{MAFA-FMD}                                                                                                  \\ \hline \hline
\textbf{Number of annotations}                & 39,485                                                                      & 56,024                                                                                                    \\ \hline
\textbf{Face type}                            & Masked face                                                                 & \begin{tabular}[c]{@{}c@{}}Unmasked face,\\ Masked face\end{tabular}                                      \\ \hline
\begin{tabular}[c]{@{}l@{}}\textbf{Mask type /} \\ \textbf{Mask wearing state}\end{tabular}                            & \begin{tabular}[c]{@{}c@{}}Simple,\\ Complex,\\ Hand,\\ Hybrid\end{tabular} & \begin{tabular}[c]{@{}c@{}}No mask wearing,\\ Correct mask wearing,\\ Incorrect mask wearing\end{tabular} \\ \hline
\textbf{Blurred face}                         & \xmark                                                                          & \cmark                                                                                                       \\ \hline
\begin{tabular}[c]{@{}l@{}}\textbf{Number of low} \\ \textbf{resolution annotations}\end{tabular} & 1,016                                                                       & 4,567                                                                                                      \\ \hline
\end{tabular}
\end{table}

\section{The MAFA-FMD Dataset}
\label{sec:MAFA-FMD}
Gu~\textit{et al.} prepared the original MAFA dataset from the Internet using the Flickr, Google, and Bing search engines~\cite{ge2017detecting}. The dataset contains 35,806 images with a minimum length of 80 pixels. The annotations of the dataset have locations of faces, mask types, etc. Each image was annotated by two individuals and verified by another. More details of MAFA can be found in~\cite{ge2017detecting}. 

However, the original MAFA annotations do not address the requirements for face mask detection in healthcare settings. Therefore, we relabelled the MAFA dataset with three different mask wearing states, ``no mask wearing'', ``correct mask wearing'', and ``incorrect mask wearing'', and named it MAFA-FMD. The procedure for relabeling is as follows. First, we generated reference annotations from original annotations. In detail, we kept all box annotations, and converted ``simple'', ``complex'' mask types as correct mask wearing, ``body'', ``hybrid'' mask types as no mask wearing states. Second, we applied RetinaFaceMask trained on the AIZOO dataset to do inference on MAFA, and recorded all predictions as another reference. Finally, three professional persons manually revised all reference box coordinates and class annotations, and used LabelImg to relabel new faces as well. When identifying masks, we considered disposable medical masks, medical surgical masks, medical protective masks, dusk masks, gas masks, respirators as valid masks. In addition, cloth masks were also regarded as valid ones, since it is also advised by the centers for disease control and prevention (CDC)~\cite{masktype}. Certain masks that do not completely enclose the mouth and nose were deemed invalid. For example, those who wear traditional Chinese veils were considered no mask wearing cases, despite the fact that they resemble some forms of masks.

The major differences between the original MAFA and the MAFA-FMD are summarized in Table~\ref{table:data_diff}. In terms of the total number of annotated faces, MAFA contains 39,485 annotated faces, while MAFA-FMD has 56,084 ones, which is around 16,000 more than that of MAFA. For face types, MAFA does not annotate faces without any masks, but MAFA-FMD contains both masked and unmasked faces. In addition, the mask types have been reclassified to mask wearing states as ``no mask wearing'', ``correct masking wearing'', ``incorrect mask wearing'' with the corresponding numbers 26,463, 28,233, and 1,388 for each class. The imbalanced label distribution shows a long-tailed problem for this in-the-wild dataset. Furthermore, MAFA-FMD includes blurred faces, which were not included in the original MAFA annotation. The number of low resolution (smaller than $32 \times 32$ resolution) annotations has been increased from approximately 1,000 in MAFA to more than 4,000 in MAFA-FMD.

\section{METHODOLOGY}
\subsection{Network Architecture}
The architecture of the proposed RetinaFaceMask is shown in Fig.~\ref{fig:RetinaFaceMask_arch}. To cope with the diverse scenes in face mask detection, a strong feature extraction network ResNet50 is used as the backbone network. $C_{1}$, $C_{2}$, $C_{3}$, $C_{4}$ and $C_{5}$ denote the intermediate output feature maps of the backbone's layers conv1, conv2\_x, conv3\_x, conv4\_x and conv5\_x used in the original ResNet50~\cite{he2016deep}. These feature maps are generated by convolutions with distinct receptive fields, allowing for the detection of objects of varying sizes. At this point, we have established the general structure for our multi-scale detection model. However, one disadvantage of shallow layers is that their outputs lack sufficient high-level semantic information, which might result in poor detection performance. To address this, an FPN has been adopted, and the details are as follows. First, we apply a $3 \times 3$ convolution on $C_{5}$ to obtain $P_{5}$. Then, we upsample $P_{5}$ using nearest interpolation to the same size as $C_{4}$, and merge the upsampled $P_{5}$ and channel-adjusted $C_{4}$ with an element-wise addition. Likewise, we obtain $P_{3}$ from $P_{4}$ and $C_{3}$. In addition, we also proposed a light-weighted version of RetinaFaceMask (RetinaFaceMask-Light) by using the backbone of MobileNetV1 for running on embedded devices efficiently. $C_{3}$, $C_{4}$ and $C_{5}$ for RetinaFaceMask-Light are yielded from the last convolution blocks with the original output sizes $28 \times 28$, $14 \times 14$, and $7 \times 7$ in~\cite{howard2017mobilenets}.

\subsection{Context Attention Module}
\begin{figure}[!tb]
    \centering
    \includegraphics[keepaspectratio=true,width=0.8\linewidth, width=20pc]{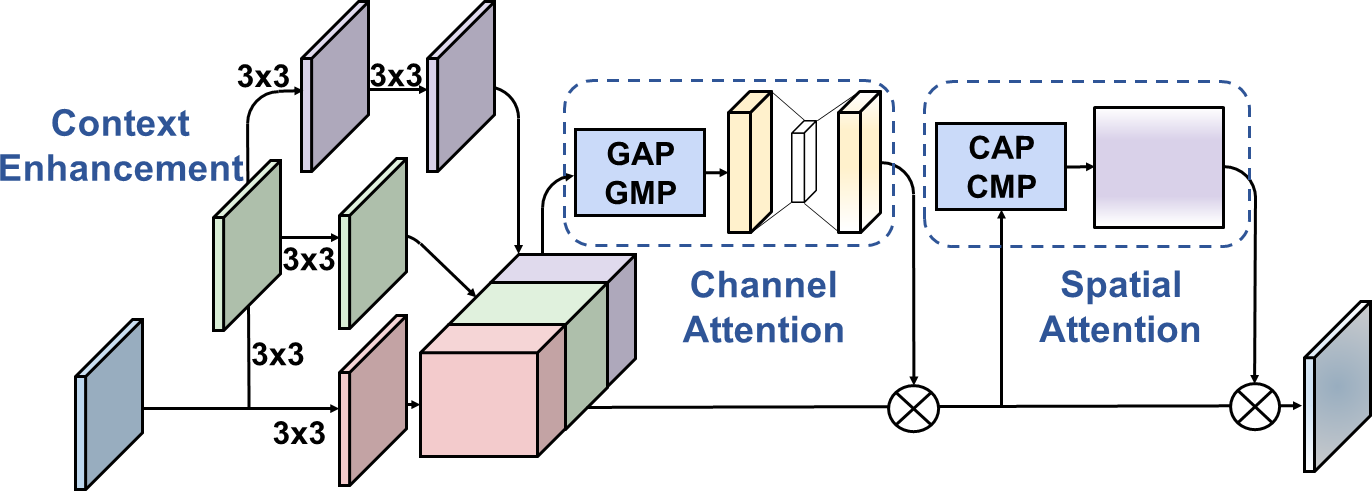}
    \caption{Illustration of CAM. It has a context enhancement block, a channel attention block, and a spatial attention block.}
    \label{fig:context_attention}
\end{figure}

In comparison to face detection, face mask detection requires both the localization of faces and the discrimination of distinct mask wearing states. To focus on learning more discriminated features for mask wearing states, we proposed a CAM as shown in Fig.~\ref{fig:context_attention}. First, to enhance the context feature extraction, we employ three parallel subbranches consisting of one $3 \times 3$ convolution, two $3 \times 3$ convolutions, and three $3 \times 3$ convolutions. Equivalently, these branches correspond to $3 \times 3$, $5 \times 5$ and $7 \times 7$ receptive fields. Then, inspired by~\cite{woo2018cbam}, we apply channel and spatial attention to focus on both channel and spatial important features associated with face mask wearing states. The channel attention block  on the input $P\in\mathbb{R}^{D\times H\times W}$ can be calculated as
\begin{equation}
\begin{adjustbox}{max width=0.44\textwidth}    
    $\Lambda_{c}=\sigma\bigg(\mathcal{F}_{{MLP}}\big(\mathcal{H}_{{GAP}}(P)\big)+\mathcal{F}_{{MLP}}\big(\mathcal{H}_{{GMP}}(P)\big)\bigg)\in\mathbb{R}^D$,
\end{adjustbox}
\end{equation}
where $\Lambda_{c}$ is the channel attention map; sigmoid function $\sigma$ normalizes the output to $(0,1)$; $\mathcal{F}_{{MLP}}$ denotes for a three-layer multi-layer perception; $\mathcal{H}_{{GAP}}$ and $\mathcal{H}_{{GMP}}$ are global average pooling and global maximum pooling. Similarly, the attention map $\Lambda_{s}$ yielded by the spatial attention block is
\begin{equation}
\begin{adjustbox}{max width=0.44\textwidth}    
    $\Lambda_{s}=\sigma\bigg(K_{3\times3}\odot\Big(\mathcal{H}_{{CAP}}(P)\oplus \mathcal{H}_{{CMP}}(P)\Big)\bigg)\in\mathbb{R}^{H\times W}$,
\end{adjustbox}
\end{equation}
where $\odot$ denotes a 2D convolution; $K_{3\times3}$ is a $3\times3$ kernel; $\oplus$ stands for the channel concatenation; $\mathcal{H}_{{CAP}}$ and $\mathcal{H}_{{CMP}}$ are channel average pooling and channel maximum pooling.

\subsection{Transfer Learning}
The uncontrolled and diverse in-the-wild scenes make feature learning difficult. One possible solution is to collect and annotate more data for training. In RetinaFaceMask, we proposed to mimic the human learning process by transferring knowledge from face detection to help face mask detection. According to~\cite{zamir2018taskonomy, fan2020hybrid}, TL has aided in feature learning as long as these tasks have a correlation. Therefore, in our work, we transfer the knowledge learned on a large scale face detection dataset Wider~Face, which consists of 32,203 images and 393,703 annotated faces~\cite{yang2016wider} to enhance the feature extraction ability for FMD. 

\subsection{Training}
\label{sc:loss}
Our network generates two matrices, location offset $\widehat{y}_{l}\in\mathbb{R}^{n_p\times4}$ and class probability $\widehat{y}_{c}\in\mathbb{R}^{n_p\times n_c}$, where $n_p$ and $n_c$ refer to the number of anchors and the number of categories of the bounding boxes, respectively. The following data, default anchors $y_{da}\in\mathbb{R}^{n_p\times4}$, the ground truth bounding boxes $y_{l}\in\mathbb{R}^{n_o\times4}$ and the true class label $y_{c}\in\mathbb{R}^{n_o\times1}$ are provided, where $n_o$ is the number of objects to be detected and is variable for different images.

To calculate the model's loss, we begin by selecting the top class and calculating the offset for each default anchor through matching the default anchors $y_{da}$, the ground truth bounding boxes $y_{l}$, and the true class label $y_{c}$ to obtain matched matrices $p_{ml}\in\mathbb{R}^{n_p\times4}$ and $p_{mc}\in\mathbb{R}^{n_p}$, where the rows of $p_{ml}$ and $p_{mc}$ denote the coordinates offsets and the labels with the highest probability for each default anchor, respectively. Then, we obtain the positive localization prediction and positive matched default anchors $\widehat{y}_{l}^+\in\mathbb{R}^{p_+\times4}$ and $p_{ml}^+\in\mathbb{R}^{p_+}$ by selecting the foreground boxes, where $p_+$ denotes the number of default anchors with non-zero top classification label. The $L_1$-smooth loss $L_{loc}(\widehat{y}_{l}^+, p_{ml}^+)$ is used to perform box coordinates regression. Following that, the hard negative mining~\cite{shrivastava2016training} is performed to obtain the sampled negative default anchors $p_{mc}^-\in\mathbb{R}^{p_-}$ and predicted anchors $\widehat{y}_{c}^-\in\mathbb{R}^{p_-}$, where $p_-$ is the number of sampled negative anchors. Finally, we calculate the classification confidence loss by $L_{conf}(\widehat{y}_{c}^-, p_{mc}^-)+L_{conf}(\widehat{y}_{c}^+, p_{mc}^+)$.

In summary, the total loss is calculated as follows,
\begin{flalign}
\hspace{-0.2cm}
\begin{adjustbox}{max width=0.45\textwidth}
    $L =\frac{1}{n_{m}}( L_{conf}(\widehat{y}_{c}^-, p_{mc}^-)+L_{conf}(\widehat{y}_{c}^+, p_{mc}^+) + \alpha L_{loc}(\widehat{y}_{l}^+, p_{ml}^+))$,    
\end{adjustbox}
\end{flalign}
where $n_m$ is the number of matched default anchors, and $\alpha$ is a weight for the localization loss.

\subsection{Inference}
In the inference stage, the trained model generates the object's localization $\widehat{y}_l\in\mathbb{R}^{n_p\times4}$ and confidence $\widehat{y}_c\in\mathbb{R}^{n_p\times4}$, where the second column of $\widehat{y}_c$ denoted as $\widehat{y}_{n}\in\mathbb{R}^{n_p}$ is the probability of no mask wearing states; the third column of $\widehat{y}_{c}$ denoted as $\widehat{y}_{cm}\in\mathbb{R}^{n_p}$ is the confidence of correct mask wearing states; the fourth column of $\widehat{y}_c$ denoted as $\widehat{y}_{im}\in\mathbb{R}^{n_p}$ is the confidence of incorrect mask wearing states. We remove objects with confidences lower than $t_c$ and perform the NMS with IoUs larger than $t_{nms}$ to obtain the final predictions.

\section{EXPERIMENT AND DISCUSSION}
\subsection{Dataset}
\subsubsection{AIZOO}
The AIZOO Face Mask Dataset~\cite{AIZOO} has 7,959 images, where the faces are annotated either with a mask or without a mask. The dataset is a composite of the Wider Face\cite{yang2016wider} and MAFA datasets~\cite{ge2017detecting}, with approximately $50\%$ of data from each. The predefined test set is used.

\subsubsection{MAFA-FMD}
As described in section \ref{sec:MAFA-FMD}, MAFA-FMD is a reannotated dataset, in which there are three classes, ``no mask wearing'', ``correct mask wearing'' and ``incorrect mask wearing''. The original test set split of MAFA is kept.

\subsection{Experiment Setup}
The model was developed on PyTorch \cite{paszke2019pytorch} deep learning framework. The model was trained for $250$ epochs with a stochastic gradient descent (SGD) algorithm of learning rate $10^{-3}$ and momentum $0.9$. An NVIDIA GeForce RTX 2080 Ti GPU was employed. The input image resolution is $840 \times 840$ for RetinaFaceMask, and is $640 \times 640$ for RetinaFaceMask-Light.

\subsection{Ablation Study}
We performed an ablation study to evaluate the effectiveness of CAM and TL using RetinaFaceMask on the AIZOO dataset. We used average precision (AP) for each class, and mean average precision (mAP) as the evaluation metrics~\cite{padilla2020survey}. AP\textsubscript{N} and AP\textsubscript{M} are APs for no mask wearing and mask wearing states, respectively. The experiment results were summarized in Table~\ref{table:ablation} and the best result was obtained by combing CAM and TL. The following sections discuss the effectiveness of each module.

\subsubsection{Context Attention Module}
By including CAM in the model, we observed an around $1\%$ increase in mAP. In particular, AP for no mask wearing increased from $92.8\%$ to $94.2\%$, and AP for mask wearing improved from $93.1\%$ to $93.6\%$. These findings indicate that CAM can be used to focus on the desired face and mask features, which can alleviate the effect of the imbalanced problem.

\subsubsection{Transfer Learning}
To evaluate the performance of TL using face detection knowledge, we added TL to the model. We noticed a considerable rise in mAP from $93.0\%$ to $94.4\%$ when compared to the baseline. The possible reason for this is because face detection and face mask detection are highly related, and so the features learned for the former become beneficial for the latter. 

\begin{table}[]
    \centering
    \vspace{0.2cm}
    \caption{Ablation Study of RetinaFaceMask.}
    \begin{tabular}{|c|c|c|c|c|}
    \hline
    \textbf{CAM}          & \textbf{TL}           & \textbf{AP\textsubscript{N}} & \textbf{AP\textsubscript{M}} & \textbf{mAP} \\ \hline \hline
    \xmark & \xmark & 92.8              & 93.1              & 93.0           \\ \hline
    \cmark & \xmark & 94.2              & 93.6              & 93.9           \\ \hline
    \xmark & \cmark & 94.5              & 94.3              & 94.4           \\ \hline
    \cmark & \cmark & 95.0              & 94.6              & 94.8           \\ \hline
    \end{tabular}
    \label{table:ablation}
    \vspace{-0.1cm}
\end{table}

\subsection{Comparison with other Methods}

\begin{figure*}[!t]
    \vspace{0.5cm}
	\centering
	\subfigure[AIZOO]{
		\includegraphics[keepaspectratio=true,width=0.85\textwidth, width=42pc]{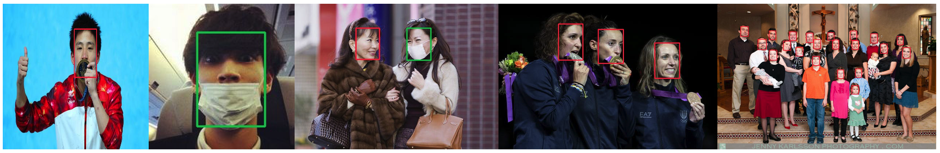}
		\label{fig:qualitative_result_aizoo}
	}\\
	\subfigure[MAFA-FMD]{
		\includegraphics[keepaspectratio=true,width=0.85\textwidth, width=42pc]{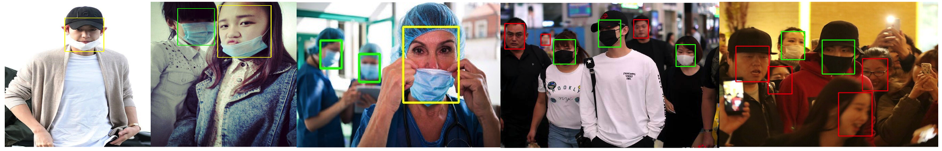}
		\label{fig:qualitative_result_mafa_fmd}
	}\\
	\caption{Qualitative Results on AIZOO and MAFA-FMD Datasets. Red boxes are no mask wearing on both datasets; green boxes are mask wearing on AIZOO, and correct mask wearing on MAFA-FMD; yellow boxes are incorrect mask wearing on MAFA-FMD.}
	\label{fig:qualitative_result}
 	\vspace{-0.1cm}
\end{figure*}

\subsubsection{Comparison on AIZOO}
In Table~\ref{table:result_aizoo}, we compared our model's performance with that of other widely used detectors for face mask detection. SSD is the baseline approach released by the AIZOO dataset's produce~\cite{AIZOO}. YOLOv3 has been used in numerous face mask detection investigations~\cite{li2020robust, ren2020mask}. RetinaFace was also included in the comparison as an efficient face detector. We discovered that RetinaFaceMask can outperform YOLOv3 and RetinaFace by $1.7\%$ and $1.8\%$, respectively, and obtain the state-of-the-art result in terms of mAP. Additionally, for the APs with and without masks, RetinaFaceMask demonstrated the best outcome. Our lite version, RetinaFaceMask-Light, which utilizes a significantly smaller model, achieved an acceptable result of $92.0\%$ in mAP. It should be noted that the number of parameters in RetinaFaceMask-Light is much less than other models.

Additionally, we showed some qualitative AIZOO dataset results in Fig~\ref{fig:qualitative_result_aizoo}. As seen in the first and fourth images, the model is robust to confusing masking types. In the second and third images, faces with mask wearing were correctly spotted. We discovered that one of an infant's little faces was omitted from the last image. One probable explanation for this is that the training dataset lacks small faces, and hence the model does not learn a good representation for such faces.

\begin{table}[t]
\centering
\caption{Comparison with other methods on AIZOO in percentage.}
\begin{tabular}{|l|c|c|c|}
\hline
\textbf{Method}               & \textbf{AP\textsubscript{N}} & \textbf{AP\textsubscript{M}} & \textbf{mAP}  \\ \hline \hline
\textbf{SSD~\cite{liu2016ssd}}                  & 89.6                                            & 91.9                                            & 90.8          \\ \hline
\textbf{Faster R-CNN~\cite{ren2015faster}}         & 83.3                                            & 83.7                                            & 83.5          \\ \hline
\textbf{YOLOv3~\cite{redmon2018yolov3}}               & 92.6                                            & 93.7                                            & 93.1          \\ \hline
\textbf{RetinaFace~\cite{deng2020retinaface}} & 92.8 & 93.1 & 93.0 \\
\hline \hline
\textbf{RetinaFaceMask}       & \textbf{95.0}                                   & \textbf{94.6}                                   & \textbf{94.8} \\ \hline
\textbf{RetinaFaceMask-Light} & 93.6                                            & 90.4                                            & 92.0          \\ \hline
\end{tabular}
\label{table:result_aizoo}
\end{table}

\subsubsection{Comparison on MAFA-FMD}
We also compared our method's performance on the MAFA-FMD dataset. Additional evaluation metrics: AP\textsubscript{CM} for the correct mask wearing, and AP\textsubscript{IM} for the incorrect mask wearing, are included. Since we only annotated masks that can protect humans in healthcare settings as valid masks, some masks which do not enclose the faces are denoted as no mask wearing. This may increase the hardness of learning, because they are hard to distinguish. In addition, the three-class task is likely to be harder than the two-class task. Although it is hard, our method still achieved the state-of-the-art performance on mAP and APs of different classes as shown in Table~\ref{table:result_mafa_fmd}. Compared to the second best method RetinaFace, we had an around $2\%$ improvement in mAP. However, our light-weighted version RetinaFaceMask-light only obtained a $59.8\%$ mAP, which may be due to the reason that light and shallow models are hard to learn enough useful features.

Fig.~\ref{fig:qualitative_result_mafa_fmd} illustrates some qualitative findings from the MAFA-FMD dataset. In comparison to the second AIZOO image in Fig.~\ref{fig:qualitative_result_aizoo}, the model trained on our reannotated dataset is capable of correctly discriminating between correct and incorrect mask wearing cases, as demonstrated by the first three images. Additionally, the MAFA-FMD trained model is capable of capturing some small or blurred faces. However, rare failures may occur when the face is occluded by someone or something.

\begin{table}[t]
\vspace{0.2cm}
\centering
\caption{Comparison with other methods on MAFA-FMD in percentage.}
\begin{tabular}{|l|c|c|c|c|}
\hline
\textbf{Method}               & \textbf{AP\textsubscript{N}} & \textbf{AP\textsubscript{CM}} & \textbf{AP\textsubscript{IM}} & \textbf{mAP}  \\ \hline \hline
\textbf{SSD~\cite{liu2016ssd}}                  & 46.5                                               & 80.7                                                & 17.7                                                & 48.3          \\ \hline
\textbf{Faster R-CNN~\cite{ren2015faster}}         & 55.7                                               & 86.3                                                & 43.9                                                & 62.0          \\ \hline
\textbf{YOLOv3~\cite{redmon2018yolov3}}               & 61.3                                            & 88.9                                             & 48.1                                             & 66.1          \\ \hline
\textbf{RetinaFace~\cite{deng2020retinaface}}           & 58.7                                            & 87.4                                             & 53.3                                             & 66.5          \\ \hline \hline
\textbf{RetinaFaceMask}       & \textbf{59.8}                                   & \textbf{89.6}                                    & \textbf{55.6}                                    & \textbf{68.3} \\ \hline
\textbf{RetinaFaceMask-Light} & 55.9                                            & 88.6                                             & 34.9                                             & 59.8          \\ \hline
\end{tabular}
\label{table:result_mafa_fmd}
\end{table}

\section{CONCLUSIONS}
In this paper, we proposed a novel single stage face mask detector, namely RetinaFaceMask. We made the following contributions. First, we created a new face mask detection dataset, MAFA-FMD, with a more realistic and informative classification of mask wearing states. Second, we proposed a new attention module, CAM, that would be dedicated to learning discriminated features associated with face mask wearing states. Third, we emulated humans' ability to transfer knowledge from the face detection task to improve face mask detection. The proposed method achieved state-of-the-art results on the public face mask dataset as well as our new dataset. In particular, compared with the baseline method on the AIZOO dataset, we have improved the mAP by $4\%$ than the baseline. Therefore, we believe our method can benefit both the emerging field of face mask detection and public healthcare to combat the spread of COVID-19. Further work may include tackling problems of occlusions or small faces in face mask detection.




\section*{ACKNOWLEDGMENT}
\noindent The authors thank Prof. H. Yan for valuable discussion.



\bibliographystyle{IEEEtran}
\bibliography{Reference_BibTex}

\end{document}